\documentclass[conference]{IEEEtran}
\IEEEoverridecommandlockouts
\usepackage{cite}
\usepackage{amsmath,amssymb,amsfonts}
\usepackage{algorithmic}
\usepackage{amsthm}
\usepackage{bbm}
\usepackage{url}
\usepackage{graphicx}
\usepackage{textcomp}
\usepackage{xcolor}
\def\method{\texttt{LHGEL}}

\newtheorem{theorem}{Theorem}

\usepackage[linesnumbered,ruled]{algorithm2e}

\newtheorem{remark}{Remark}
\def\rg{\text{relation group}}
\def\BibTeX{{\rm B\kern-.05em{\sc i\kern-.025em b}\kern-.08em
    T\kern-.1667em\lower.7ex\hbox{E}\kern-.125emX}}
\begin{document}

\title{LHGEL: Large Heterogeneous Graph Ensemble Learning using Batch View Aggregation}

\author
{\IEEEauthorblockN{Jiajun Shen\IEEEauthorrefmark{1}, Yufei Jin\IEEEauthorrefmark{1}, Yi He\IEEEauthorrefmark{2}, and Xingquan Zhu\IEEEauthorrefmark{1}}
\IEEEauthorblockA{\IEEEauthorrefmark{1}{Dept. of Electrical Engineering and Computer Science, Florida Atlantic University 
Boca Raton, Florida, USA}} 
\IEEEauthorblockA{\IEEEauthorrefmark{2}{Department of Data Science, William \& Mary, Williamsburg, Virginia, USA}} 
jshen2024@fau.edu, yjin2021@fau.edu, yihe@wm.edu, xzhu3@fau.edu
}

\maketitle

\begin{abstract} Learning from large heterogeneous graphs presents significant challenges due to the scale of networks, heterogeneity in node and edge types, variations in nodal features, and complex local neighborhood structures. 
%
%
This paper advocates for ensemble learning as a natural solution to this problem, whereby training multiple graph learners under distinct sampling conditions, the ensemble inherently captures different aspects of graph heterogeneity. 
Yet, the crux lies in combining these learners to meet global optimization objective while maintaining computational efficiency on large-scale graphs.
%
%
In response, we propose~\method, an ensemble framework that addresses these challenges through batch sampling with three key components, namely \textit{batch view aggregation}, \textit{residual attention}, and  \textit{diversity regularization}. 
Specifically, batch view aggregation samples subgraphs and forms multiple graph views, while residual attention adaptively weights the contributions of these views to guide node embeddings toward informative subgraphs, thereby improving the accuracy of base learners.
Diversity regularization 
encourages representational disparity across embedding matrices derived from different views, promoting model diversity and ensemble robustness.
Our theoretical study demonstrates that residual attention mitigates gradient vanishing issues commonly faced in ensemble learning. 
Empirical results on five real heterogeneous networks validate that our \method\ approach consistently outperforms its state-of-the-art competitors by substantial margin. 
Codes and datasets are available at \url{https://github.com/Chrisshen12/LHGEL}.
%
%

\end{abstract}


\section{Introduction}






\noindent
Ensemble learning strives to combine predictions from multiple base learners
to improve model accuracy and robustness.
Whereas its effectiveness has been widely documented in classical machine learning~\cite{Ensemble2018sagi} and modern deep learning pipelines~\cite{Ensemble2022Ganaie},
most existing ensemble frameworks are developed under the assumption that data are independent and identically distributed (IID). 
%
%
%
This assumption does not hold in many real-world scenarios involving graph-structured data, such as social networks~\cite{mag}, citation networks~\cite{citation}, and urban infrastructure networks~\cite{urban}, which are inherently non-IID due to complex and structured dependencies between entities.
These graphs are often large-scale and heterogeneous, comprising multiple types of nodes, edges, and relational patterns. Despite the importance of learning from such graphs, the use of ensemble learning in this domain remains underexplored, particularly for large heterogeneous graphs.
%

Applying ensemble learning to large heterogeneous graphs presents several  unique challenges. First, 
%
%
heterogeneity in node and edge types necessitates models that can effectively integrate multi-relational semantics, such as meta-path-based representations~\cite{mag} or attention mechanisms~\cite{HAN}.
Second, the sheer size of real-world graphs demands efficient and scalable learning strategies. 
%
%
Third, training multiple base learners that are both accurate and diverse is complicated by the shared and interconnected nature of graph neighborhoods, which can lead to correlated errors and reduce ensemble effectiveness.
%

To address these challenges, this paper proposes a novel ensemble learning framework tailored for large heterogeneous graphs. Our method, named \method, adopts a batch-based sampling strategy, where 
each base learner operates on a distinct subgraph induced by a subset of sampled nodes from the input graph. 
This allows scalable and memory-efficient graph training, as the subgraph sizes can be easily controlled~\cite{batch1}. Moreover, empirical studies suggest that mini-batch training converges faster than full-graph training across various datasets, and often achieves comparable or even superior  accuracy in various settings~\cite{batch3}. 
Nevertheless, 
a known limitation of batch-based approaches is their higher variance across training runs, partly due to the stochastic nature of sampling~\cite{batch2}.
Such variance arises from sampling bias, where each batch captures only a localized view of the graph that may differ substantially across batches~\cite{batch3}. Addressing this challenge requires new mechanisms that stabilize learning while preserving the benefits of batch sampling.

To tackle these challenges, our \method\ framework incorporates three key components: (1) \textit{batch view aggregation}, which facilitates information sharing across multiple sampled subgraphs, enabling scalable training without compromising global context; 
(2) a \textit{residual-attention} mechanism, which  effectively integrates  embeddings from base learners by emphasizing informative subgraph views; and 
(3) a \textit{diversity-regularization} term,  which explicitly encourages the training of accurate  yet  complementary base learners. 
Intuitively, the batch-wise processing enables scaling to large graph volumes, while the aggregation and attention modules mitigate the local bias introduced by subgraph sampling. 
%
%
They together form an ensemble learning framework that is efficient and robust to structural and semantic variations in heterogeneous graphs.

\smallskip\noindent
{\bf Specific contributions} of this paper are as follows:
\begin{itemize}
    \item We propose a novel ensemble learning framework \method\ tailored for large heterogeneous graphs.
    \item We introduce batch view aggregation to enable information sharing across independently sampled subgraphs.
    \item We conform residual attention to integrate base learner embeddings by adaptively weighting graph views.
    \item We design a diversity regularization term to encourage the learning of diverse and accurate graph base classifiers.
\end{itemize}


\section{Related Work}
\subsection{Heterogeneous Graph Neural Networks}

Several Graph Neural Network (GNN) architectures have been proposed to handle heterogeneous graphs. \textit{RGCN} (Relational Graph Convolutional Network) \cite{RGCN} explicitly models different types of relations (edges) using relation-specific transformations and aggregations. \textit{HAN} (Heterogeneous Attention Network) \cite{HAN} introduces a hierarchical attention mechanism to learn the importance of different node types and meta-paths. \textit{MAGNN} (Meta-Aggregation in Heterogeneous Graph Neural Networks) \cite{mag} proposes a meta-path-based aggregation scheme that captures high-order semantic relationships.

Although these heterogeneous GNNs excel in capturing the complexity of heterogeneous graphs, scaling them to large graphs remains a challenge. Combining these architectural advances with sampling or clustering techniques is a promising direction to address scalability in large heterogeneous graphs.

\subsection{Graph Ensemble Learning}
Early approaches, such as graph representation ensembling~\cite{ensemble}, combines node embeddings from multiple models using simple concatenation. Building on this idea, stacking-based frameworks~\cite{ensemble2} introduces multi-level classifiers to integrate representations for tasks such as link prediction. Advanced methods, such as Graph Ensemble Neural Network (GEN)~\cite{ensemble3}, embeds ensemble learning into the GNN training process, rather than applying it only to the inference stage.

To enhance robustness and mitigate overfitting, 
recent developments have proposed more sophisticated ensemble strategies. For example, Graph Ensemble Learning (GEL)~\cite{ensemble4} incorporates serialized knowledge transfer and multilayer DropNode techniques to encourage diversity among models. Similarly, GNN-Ensemble~\cite{ensemble5} leverages substructure-based training to improve resilience against adversarial perturbations. HGEN~\cite{shen2025hgen} addresses ensemble learning for heterogeneous graphs by combining meta-path and transformation-based optimization pipeline to uplift classification accuracy. 

\subsection{Heterogeneous Graph Scaling}
Scaling GNNs to large graphs has been a significant area of research. Traditional GNNs often suffer from the neighborhood explosion problem, where the number of nodes involved in the computation grows exponentially with the number of layers. This issue is exacerbated in large and complex heterogeneous graphs.
To mitigate the neighborhood explosion problem, various sampling techniques have been proposed. \textit{GraphSAGE} \cite{SAGE} learns node embeddings by sampling and aggregating features from a fixed-size neighborhood. This approach allows generalization to unseen nodes and significantly reduces the computational cost compared to full-graph methods. \textit{FastGCN} \cite{fastgcn} reformulates the loss and gradients over random walks and uses importance sampling to reduce the variance. While these sampling based methods can be applied to heterogeneous graphs by treating different node and edge types as distinct features, it doesn't explicitly model the rich semantic relationships captured by different edge types. 

Clustering-based approaches aim to divide a large graph into smaller, more manageable clusters. \textit{Cluster-GNN} \cite{cluster} partitions the graph using a graph clustering algorithm (e.g., Metis) and then trains a GNN by performing mini-batch updates on these clusters to reduce memory and computational costs. 
This method focuses primarily on homogeneous graphs, and their direct application to heterogeneous graphs might not fully leverage the type-specific information.

Our approach utilizes metapaths to define distinct groups of typed relations within heterogeneous graphs, enabling the extraction of semantically meaningful substructures. For each relation group, we perform sampling, using multiple batch sizes, to capture diverse information at different scales. Embeddings generated from these different batch sizes are then ensembled through a residual attention mechanism. This multi-batch ensemble across relation groups effectively models the complex heterogeneity of the graph data.

\begin{figure*}[htbp]
    \centering
    \includegraphics[width=0.8\textwidth]{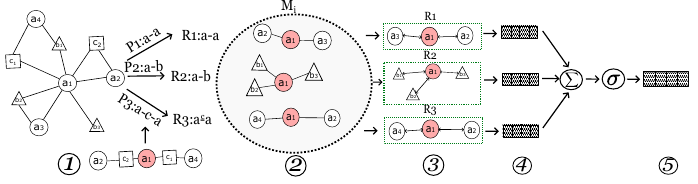}
    \caption{A conceptual view of heterogeneous information aggregation via Relational Aggregation for a target node $a_1$ (\textit{i.e.} node colored in red circle). From left to right, \textcircled{1} is the large heterogeneous graph; \textcircled{2}: a relation group $\mathcal{M}_i$ consists of node relations derived from metapaths; \textcircled{3} and \textcircled{4}: for each $\mathcal{M}_i$, we aggregate information across its relations by propagating messages along the associated relations, enabling the model to capture rich semantic dependencies as formulated in Eq.~(4); and \textcircled{5} final representation for the target node (\textit{i.e.} $a_1$), aggregated from all relations.}
    \label{fig:relationAgg}
\end{figure*}
\section{Problem Definition}

\noindent
Consider a heterogeneous graph \( G = (V, E, X) \) consisting of multiple node types and edge types. $X$ denotes feature matrix for nodes with different node types. Let \(\mathcal{T}^v\) and \(\mathcal{T}^e\) denote the sets of node types and edge types, respectively. Each node \(v \in V\) is associated with a type \(\phi(v) \in \mathcal{T}^v\), and each edge $\{e = (u,v)\} \in E$ is associated with a type \(\varphi(e) \in \mathcal{T}^e\). A  meta-path is a relational sequence $(e_{i,j}e_{j,k}\ldots e_{k,j})$. Given a specific meta-path $\mathcal{P}_i$, we extract the set of typed relations
\[
\mathcal{R}_i = \left\{ (\phi(v), \varphi((v,u)), \phi(u)) \right\}
\]
traversed in each meta-path $\mathcal{P}_i$, where $\phi$ and $\varphi$ denote the node-type and edge-type mappings, respectively. For example, in Fig.~\ref{fig:relationAgg}, we extract the relation $\mathcal{R}_3:a-a$ with edge-type $c$ from the meta-path $\mathcal{P}_3:a-c-a$.

Given a specific relation $\mathcal{R}_{i}$ with \( n_t \) as the number of nodes receiving messages and \( n_s \) as the number of nodes sending messages, we can construct a heterogeneous graph, regarding only the relation $\mathcal{R}_{i}$, equipped with an adjacency matrix $A_{R_i} \in \mathbbm{1}^{n_t \times n_s}$ where $A_{R_i}[j, k]=1$ $\iff$ $\exists~r = (\phi(v_{j}),\dots, \phi(u_{k}))$ such that the edges of the relation $r$ follow $\mathcal{R}_i$, formally presented as follows.
\begin{equation}
\begin{aligned}
\text{Gen}(G, \mathcal{R}) &= A_{R_i} \quad \text{where}\\ 
A_{R_i}[j,k] &=
\begin{cases} 
1 & \text{if } \exists r = (\phi(v_{j}),\dots, \phi(u_{k})) \text{ matching } \mathcal{R}_i, \\
0 & \text{otherwise}.
\end{cases}
\end{aligned} \label{eq:path}
\end{equation}

For example, for relation $\mathcal{R}_3$ in Fig.~\ref{fig:relationAgg}, we can obtain an adjacency matrix for node type $a$ and $a$ for a target node $a_1$: 
\[
\begin{array}{c|cccc}
    & \text{$a_1$} & \text{$a_2$} & \text{$a_3$} & \text{$a_4$}\\
\hline
\text{$a_1$} & 0 & 1 & 0 & 1\\
\text{$a_2$} & 1 & 0 & 0 & 0\\
\text{$a_3$} & 0 & 0 & 0 & 0\\
\text{$a_4$} & 1 & 0 & 0 & 0\\
\end{array}
\]
To capture diverse semantic perspectives, different combinations of relations are grouped into sets. Each unique set, called a \rg, is denoted by $\mathcal{M}_i$. The \rg~is the set of different relations:
For each \rg~$\mathcal{M}_i$, $m$ is the number of relations in $\mathcal{M}_i$ (which may vary). 
The collection of $\mathcal{M}_i$ is denoted by $\mathcal{Q}=\{\mathcal{M}_1, \ldots, \mathcal{M}_c\}$ with $c$ indicating the number of \rg~used. Additionally, a set of batch sizes $\mathcal{B} = \{b_1, \ldots, b_m\}$ is leveraged to diversify the extent of neighborhood sampling and propagation depth. Independent base graph learners are then trained on each batch size $b\in\mathcal{B}$ and each \rg~$\mathcal{M}_i\in\mathcal{Q}$. By utilizing different neighbor hops and unique relation combination, the ensemble can exploit structural and semantic variability.
Our \textbf{goal} is to enable ensemble learning of multiple base learners to predict the label for the target nodes and maximize the classification accuracy.

\begin{figure*}[htbp]
    \centering
    \includegraphics[width=0.7\textwidth]{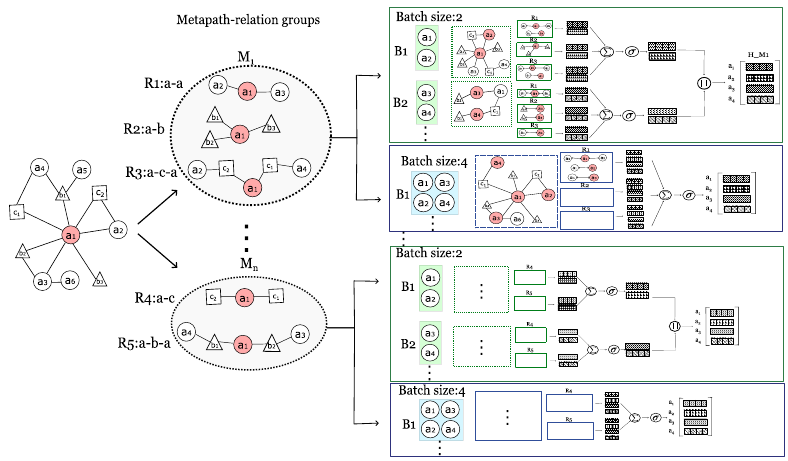}
    \caption{A conceptual overview of heterogeneous information aggregation enabled by the Batch View Aggregation mechanism. From left to right, \textcircled{1} is the large heterogeneous graph; \textcircled{2}: we extract multiple relation groups from the large heterogeneous graph; \textcircled{3}: for each relation group $\mathcal{M}_{i}$, we randomly sample a small batch of target nodes and expand their neighborhoods; \textcircled{4}: for each batch, we obtain node embeddings through relational aggregation as illustrated in Fig.~\ref{fig:relationAgg}; and \textcircled{5} final representation of target nodes at each batch size}
    \label{fig:relation}
\end{figure*}

\section{Proposed Framework}
Our model is primarily motivated by Relational Graph Convolutional Network (RGCN)~\cite{RGCN}, which handles multi-relational heterogeneous graphs by learning relation-specific transformations. In RGCN, message passing is performed separately for each relation type, and the resulting messages are aggregated to update node representations. This allows RGCN to capture semantics of diverse edge types, making it well-suited for heterogeneous graphs. However, RGCN suffers from two major limitations: (1) it does not explicitly encourage diversity among relation-specific encoders, and (2) it is limited in its ability to capture high-order semantics, as it only considers one-hop direct relations and does not support metapath-based multi-hop connectivity.
In contrast, our framework constructs relation groups based on metapaths, which represent meaningful multi-hop patterns (\textit{e.g.}, Author–Paper–Author). This enables \method~ to encode richer contextual information that goes beyond immediate neighbors.

Specifically, \method~ integrates three core components to simultaneously enhance both the accuracy and diversity of base learners: (1) a batch view aggregation mechanism to promote interaction between batches, (2) a residual-attention module to adaptively combine base learner outputs, and (3) a regularization term to encourage diversity. This design allows \method~ to generalize the relational modeling power of RGCN while overcoming its expressiveness limitations through ensemble learning.


\subsection{Relational Aggregation}
To compute node representations, we first apply Dropout to the input features and perform a linear transformation:
\begin{align}
    \tilde{X}_i~~ &= \text{Dropout}(X_i, p) \\
    H_i^{(0)} &= \sigma(\tilde{X}_i W_i^{(0)})
\end{align}

where, \(X_i\) is the input feature matrix for all nodes in \rg~$\mathcal{M}_{i}$. \(\text{Dropout}(X_i, p)\) randomly zeroes elements of \(X_i\) with probability \(p\). \(W_i^{(0)}\) is a learnable weight matrix, and \(\sigma(\cdot)\) is a nonlinear activation function.


For all nodes in \rg~$\mathcal{M}_{i}$, the simple message passing framework at layer \(l\) is computed as:
\begin{align}
    H_{\mathcal{M}_{i}}^{(l+1)} = \sigma(\sum_{\mathcal{R}_{j}\in\mathcal{M}_{i}}\tilde{A}_{\mathcal{R}_{j}}H^{(l)}_{\mathcal{R}_{j}}W_{\mathcal{R}_{j}}^{(l)})
\end{align}
where $H^{(l)}_{\mathcal{R}_{j}}\in \mathbb{R}^{n_s \times d^{(l)}}$ is the hidden representation of nodes at layer $l$ aggregated using only relation $\mathcal{R}_{j}$. $\mathcal{R}_j$ is a relation in $\mathcal{M}_{i}$. $\tilde{A}_{\mathcal{R}_{j}}\in\mathbb{R}^{n_t \times n_s}$ is the normalized adjacency matrix, derived from $A_{R_j}$. $W_{\mathcal{R}_{j}}^{(l)}$ is the learnable weight for relation $\mathcal{R}_{j}$ at layer $l$, and $\sigma(\cdot)$ is a nonlinear activation function such as ReLU.

\begin{figure*}[htbp]
    \centering
    \includegraphics[width=0.78\textwidth]{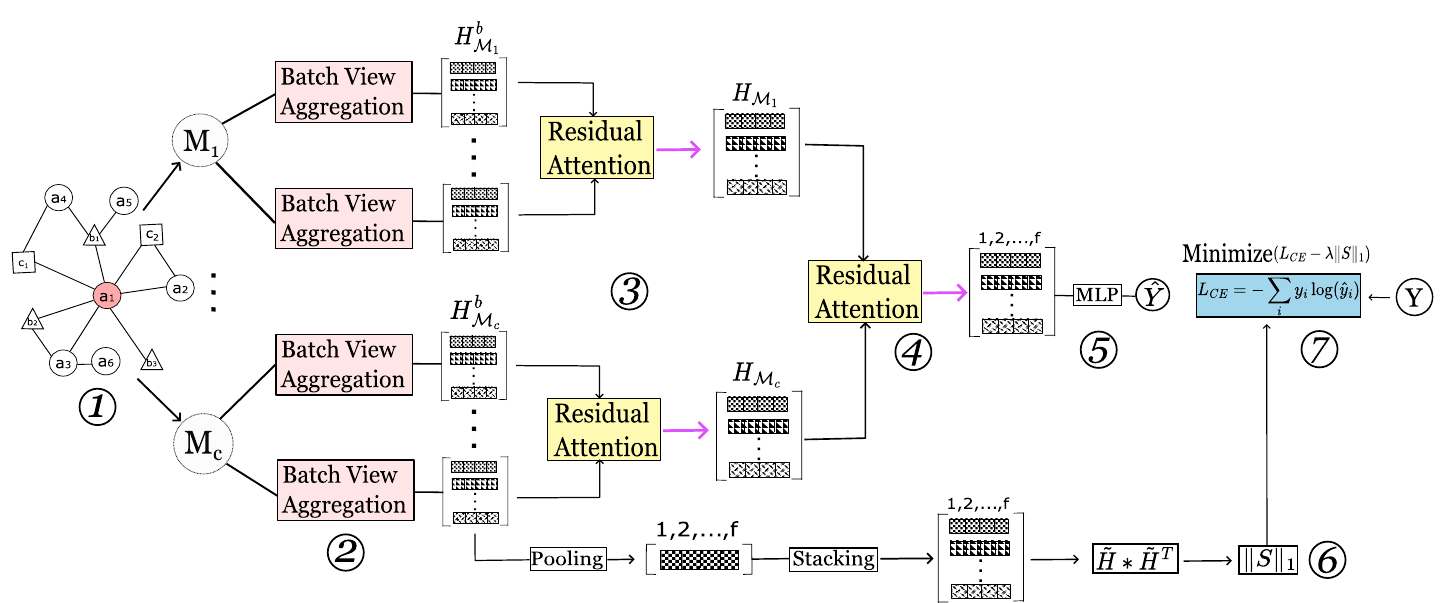}
    \caption{The proposed \method\ framework enables ensemble learning on large heterogeneous graphs. From left to right, \textcircled{1} is the large heterogeneous graph. \textcircled{2}: for each relation group, node embeddings are computed via batch aggregation as illustrated in Fig.\ref{fig:relation}. \textcircled{3}: these embeddings are then combined within each relation group using a residual attention mechanism (Fig.\ref{fig:attention-detail}), followed by a second attention-based fusion \textcircled{4} across all relation groups to produce the final node representation. \textcircled{5}: this representation is passed through a multi-layer perceptron (MLP) for prediction. \textcircled{6}: an $L_1$ norm on the correlation matrix obtained from batch aggregation embeddings intends to encourage diversity between embeddings. \textcircled{7}: The objective function jointly optimizes the batch aggregation modules, residual attention modules, and the MLP layer to achieve effective ensemble learning.}
    \label{fig:framework}
\end{figure*}

\subsection{Batch View Aggregation}

In order to scale to large heterogeneous graphs, we adopt mini-batch training to enable learning on subgraphs sampled from the full graph, which can significantly reduce memory consumption for effective training. Nevertheless, batch-based learning also imposes critical challenges. First, it relies on neighborhood sampling, leading to incomplete coverage of full graph information. Second, different batch sizes result in different levels of information propagation, with larger batches capturing broader neighborhoods, whereas smaller batches focusing on local structures. This results in different semantic views of the graph.

To address above challenges, we propose to use batch view aggregation to aggregate information for each target node, using views generated from different batch sizes. This allows \method~ to integrate both shallow and deep structure information, enhancing robustness and generalization.

\begin{enumerate}
    \item \textbf{Node Sampling:} Given a target node type $t_{\iota}\in\mathcal{T}^v$, with $V_{t_{\iota}} \subset V$ being the target node set, randomly sample a small batch of target nodes \(S \subseteq V_{t_{\iota}}\).
    \item \textbf{Neighborhood Expansion:} For each target node in \(S\), recursively expand its neighborhood by including adjacent nodes connected via \(\mathcal{R}_j\) in the relation group \(\mathcal{M}_i\), up to a predefined number of hops or size limit.
\end{enumerate}


This procedure generates a subgraph \(G_i^{b}\) corresponding to \rg~\(\mathcal{M}_i\) and batch size \(b\), which preserves the typed connectivity patterns relevant to \(\mathcal{R}_j\in \mathcal{M}_i\). Then, for each \(G_i^{b}\), we grow a subgraph \(g_k^{b}\) \textit{w.r.t.} target nodes in each batch with index $k$.
For example, in Figure~\ref{fig:relation}, for batch size 2, \textcircled{3} shows the expanded subgraph \(g_1^{2}\) and \(g_2^{2}\) from randomly sampled target nodes \textit{w.r.t.} each batch. 
For a subgraph \(g_k^{b}\), the representation for target nodes with batch size $b$ and batch index $k$ is the following:
\begin{align}
H_{\mathcal{M}_{i},k}^{(l+1), b} = \sigma \left( \sum_{\mathcal{R}_j \in \mathcal{M}_i} \tilde{A}_{\mathcal{R}_{j},k}^{b} H^{(l),b}_{\mathcal{R}_{j},k} W_{\mathcal{R}_j}^{(l)} \right)
\end{align}
\begin{align}
H_{\mathcal{M}_{i},k}^{b} = H_{\mathcal{M}_{i},k}^{(L), b}
\end{align}
where:
\begin{itemize}
    \item \( H^{(l),b}_{\mathcal{R}_{j},k} \in \mathbb{R}^{n_s \times d^{(l)}} \) is node features for source nodes with each relation $\mathcal{R}_j$ at layer $l$ with batch size $b$ at batch index $k$.
    \item \( \tilde{A}_{\mathcal{R}_{j},k}^{b} \in \mathbb{R}^{n_t \times n_s} \) is the normalized adjacency matrix for \(g_k^{b}\) via relation \( \mathcal{R}_j \). \(g_k^{b}\) is the subgraph generated from \(G_i^{b}\) with batch index $k$ and batch size $b$.
    \item \( W_{\mathcal{R}_j}^{(l)} \) is the learnable weight for relation \( \mathcal{R}_j \) at layer $l$.
    \item \( H_{\mathcal{M}_{i},k}^{b} \)is the final embedding for the target nodes in the $k_{th}$ batch for the relation group $\mathcal{M}_{i}$ and batch size $b$.
\end{itemize}

This relation-aware aggregation enables the model to capture diverse types of neighborhood information while preserving the inherent heterogeneity of the graph structure.

For different batches within each batch size group, we align node embeddings by concatenating the embeddings from batches. Denote total number of batches as $u$, we column-wise concatenate all batches with same batch size as:
\begin{align}
    H_{\mathcal{M}_{i}}^{b} = \|_{k=1}^{u} H_{\mathcal{M}_{i},k}^{b}
\end{align}

\begin{figure}[htbp]
    \centering
    \includegraphics[width=1\linewidth]{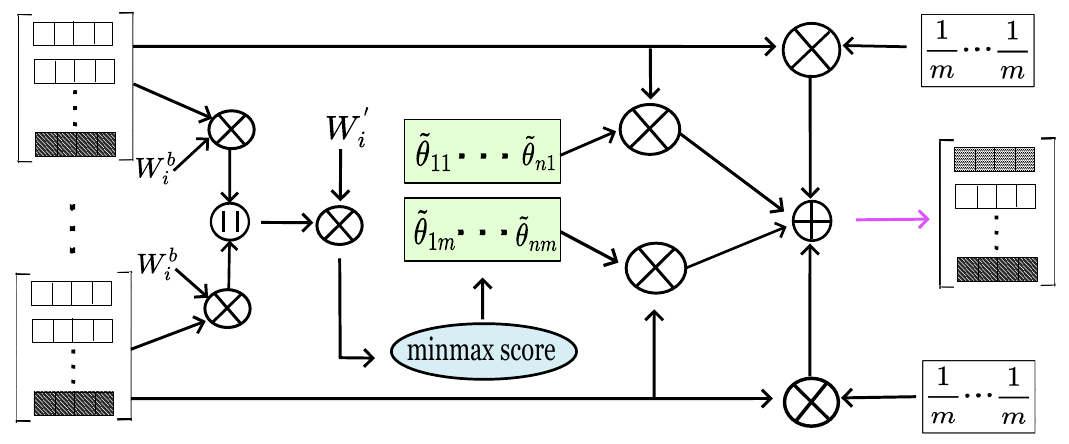}
    \caption{Residual attention computation inside each relation group. Embeddings from different batch sizes in each relation group $\mathcal{M}_{i}$ are first projected into a shared attention space through projection weight $W_{i}^b$. The projected embeddings are then concatenated and passed through a shared projection weight $W'_{i}$ to learn $m$ raw $\Theta$ attention scores. Using minmax normalization (refer to Eq. 10), the final fusion residual attention $\tilde{\Theta}_{\mathcal{M}_{i}}^{\mathcal{B}}$ is obtained. For Residual attention computation across relation groups, $W_{i}^b$ is replaced with $W_{i}$ and $m$ is replaced with $c$. The final residual attention $\tilde{\Theta}_{\mathcal{Q}}$ is obtained from Eq. 10 similarly.}
    \label{fig:attention-detail}
\end{figure}

\subsection{Residual Attention}

To aggregate node representations obtained under varying batch sizes and relation groups, we introduce a residual attention mechanism grounded in min-max normalization. This mechanism adaptively balances contributions from different submodels while preserving their unique characteristics.


We apply the residual attention mechanism in two stages. In the first stage (Eq.~\ref{eq:theta_m}), it is applied across different batch sizes within each relation group. In the second stage (Eq.~\ref{eq:theta_q}), it is applied across different relation groups to obtain the final attention scores.
\begin{equation}
\Theta_{\mathcal{M}_{i}}^{\mathcal{B}} = \big\|_{b\in \mathcal{B}}
 \left\{ H_{\mathcal{M}_{i}}^{b} W_{i}^{b} \right\}
 W_i^{'}
\label{eq:theta_m}
\end{equation}
\begin{equation}
\Theta_{\mathcal{Q}} = \big\|_{i=1}^{c}
 \left\{ H_{\mathcal{M}_{i}} W_{i} \right\}
 W_i^{'}
\label{eq:theta_q}
\end{equation}
\begin{equation}
\bar{\Theta} = \text{Mean}(\Theta), \quad \hat{\Theta} = \Theta-\bar{\Theta}, \quad
\tilde{\Theta} =
\frac{(\hat{\Theta}-\hat{\Theta}^{\downarrow})} {(\hat{\Theta}^{\uparrow}-\hat{\Theta}^{\downarrow})}
\label{eq:attention_org}
\end{equation}
where \( \hat{\Theta}_i^{\uparrow} \) and \( \hat{\Theta}_i^{\downarrow} \) denote the maximum and minimum of \( \hat{\Theta}_i \) along each feature dimension, respectively.

The final fused representation for \rg~\( \mathcal{M}_{i} \) is computed as a weighted sum over its \( m \) batch size specific representations:
\begin{align}
H_{\mathcal{M}_{i}} = \sum_{j=1}^{m} \left( \tilde{\Theta}_{\mathcal{M}_{i}}^{\mathcal{B}}[:, j] + \frac{1}{m} \right)  H_{\mathcal{M}_{i}}^{b} \label{eq:att1}
\end{align}

Finally, we use a second-level aggregation which ensures that each relation group $\mathcal{M}_{i}$ contributes adaptively based on its residual importance.
\begin{equation}
    H^{\text{final}} = \sum_{i=1}^{c} \left( \tilde{\Theta}_{\mathcal{Q}}[:, i] + \frac{1}{c} \right)H_{\mathcal{M}_{i}} \label{eq:att_final}
\end{equation}

where \( H^{\text{final}} \) is the unified representation used for downstream prediction.

\textcolor{black}{The motivation of the residual attention in Eqs. (\ref{eq:att1}) and (\ref{eq:att_final}) is to allow each relation group and each batch size to participate in the learning of the final aggregations in heterogeneous graphs, without being biased or dominated by few channels. Averaging the residual weights ensures that each relation group contributes to the output, enabling gradients to flow back through all relation types and across the batch during training. This mechanism avoids the case where initial attention scores, $i.e.,$ $\tilde{\Theta}_{\mathcal{M}_{i}}^{\mathcal{B}}$ and $\tilde{\Theta}_{\mathcal{Q}}$, may be too sparse to effectively updates from all information channels, leading to biased model dominated by a few relation groups and batch sizes.}

\subsection{Diversity Regularization}

To promote complementary information learning across different relation groups and batch settings, we introduce a diversity regularization mechanism that penalizes redundancy among predictions derived from distinct \rg~with varying batch sizes.

After obtaining the embeddings from all relation groups across multiple batch sizes, we apply mean pooling over the node dimension to generate a compact prediction vector. These vectors are then stacked to construct a matrix \( \tilde{H} \in \mathbb{R}^{n \times h} \), where \( n \) denotes the total number of nodes and \( h \) represents the embedding dimension.

To quantify the redundancy among these predictions, we compute a correlation matrix:

\begin{equation}
    S = \tilde{H} \tilde{H}^\top
\end{equation}

where \( S \in \mathbb{R}^{n \times n} \) captures pairwise linear correlations among all predictions. A high off-diagonal value in \( S \) implies strong similarity between two prediction vectors, indicating redundant information.

To encourage diversity, an \( \ell_1 \) norm penalty is imposed to the correlation matrix \( S \). This encourages sparsity in \( S \), effectively reducing overlap among the learned representations. The final training objective integrates both the classification loss and the diversity regularization term:

\begin{equation}
    \mathcal{L} = - \sum_{i} y_i \log(\hat{y}_i) + \lambda \| S \|_1
\end{equation}

This framework enables scalable node classification by leveraging partial relational views from heterogeneous graphs.


\subsection{LHGEL Algorithm}
Algorithm 1 outlines major steps of the proposed \method~for ensemble learning from large heterogeneous graphs. 
\newcommand\mycommfont[1]{\textcolor{blue}{#1}}
\SetCommentSty{mycommfont}
\begin{algorithm}[htbp]
    \small
    \SetKwInOut{Init}{Initialize}\SetKwInOut{Input}{Parameters}

    \caption{\footnotesize{\method\ Heterogeneous Graph Ensemble Learning}}
    \Init{Heterogeneous graph $G = (V, E)$,\\
         Node features $\{ X \}$,\\
         Relation group set $\{\mathcal{M}_1, \ldots, \mathcal{M}_c\}$,\\
         Batch size set $\{b_1, \ldots, b_m\}$.}

    \Input{Number of batch sizes $m$,\\Number of relation groups $c$.}
    \BlankLine

    \For{each \rg~$\mathcal{M}_{i}$ \textbf{in} $\{\mathcal{M}_{1}, \mathcal{M}_{2}, \dots, \mathcal{M}_{c}\}$}{

        \For{each batch size $b$ \textbf{in} $\{b_1, \ldots, b_m\}$}{
            $\tilde{X}_{i} \leftarrow \text{Dropout}(X_i,p)$\;
            $H_{\mathcal{M}_{i}}^{b} \leftarrow \text{Obtain embedding (Eq.~7)}$\;
            
        }
        $\tilde{\Theta}_{\mathcal{M}_{i}}^{\mathcal{B}}\leftarrow \text{Calculate normalized attentions (Eq.~10)}$\;
            \tcc{Fuse node embeddings for $\mathcal{M}_i$ over $k$ batch sizes}
            $H_{\mathcal{M}_{i}} \leftarrow \sum_{j=1}^{m} \left( \tilde{\Theta}_{\mathcal{M}_{i}}^{\mathcal{B}}[:, j] + \frac{1}{m} \right)  H_{\mathcal{M}_{i}}^{b}$\;
    }
     $\tilde{\Theta}_{\mathcal{Q}} \leftarrow \text{Calculate normalized attentions (Eq.~10)}$\;
        \tcc{Fuse node embeddings over $c$ relation groups}
    $H^{\text{final}} \leftarrow \sum_{i=1}^{c} \left( \tilde{\Theta}_{\mathcal{Q}}[:, i] + \frac{1}{c} \right)H_{\mathcal{M}_{i}}$\;
    $\hat{Y} \leftarrow \text{MLP}(H^{\text{final}})$ \tcp*{Ensemble prediction}
    $\tilde{H}_{i} \leftarrow \text{MP}(H_{\mathcal{M}_{i}}^{b})$ \tcp*{Pooling over nodes of $H_{\mathcal{M}_{i}}^{b}$}
    $\tilde{H} \leftarrow \cup_{j=1}^{n} \tilde{H}_{j}$ \tcp*{Stack $\tilde{H}_{j}$ to obtain $\tilde{H}$}
    $S \leftarrow \tilde{H} \cdot \tilde{H}^T$ \tcp*{Inter-correlation among relation groups}
    \tcc{Calculate cross-entropy loss with $L1$ norm on $S$}
    $L \leftarrow - \sum_{i} y_i \log(\hat{y}_i) + \lambda \| S \|_1$\;
    Back propagation and update parameters of \method\;
    \Return{$\hat{Y}$}
\end{algorithm}

\section{Theoretical Analysis}
\subsection{Alleviate Gradient Vanishing Problem}
In order to aggregate representations with different batch sizes and relation groups, \method~ employs a residual connection with a mean coefficient. For $k$ number of base models, the mean coefficient is $\frac{1}{k}$. In the following, we prove that this helps alleviate the gradient vanishing problem for graph ensemble learning.

For simplicity, we analyze vector form with single node case, matrix generalization over multiple nodes is easily extended. Since the aggregation method for both batch size and relation group is the same, a general index i is applied to denote different representations over multiple sources. Specifically, a single node representation $h_{i}$ is denoted as node representation learned through any message passing framework from a unique source, \textit{i.e.}\ unique combination of batch size $b$ and relation group $\mathcal{M}$.
The aggregation over all unique sources for the node representation $h_{i}$ then follows Eq.~(\ref{eq:theta_q}) as:
\begin{align}
    h_{f} &= \sum_{i}(\tilde{\theta}_{i}+\frac{1}{k})h_{i} \\
    \tilde{\theta} &= \frac{\theta-\theta^{\downarrow}}{\theta^{\uparrow}-\theta^{\downarrow}}\\
    \theta &= \|_{i}\{h_{i}W_{i}\}W^{'}\\
    W' &= 
    \begin{bmatrix}
    W'_{11} & W'_{12} & \cdots & W'_{1k} \\
    W'_{21} & W'_{22} & \cdots & W'_{2k} \\
    \vdots & \vdots & \ddots & \vdots \\
    W'_{k1} & W'_{k2} & \cdots & W'_{kk}
    \end{bmatrix},
    \quad W'_{ij} \in \mathbb{R}^{d' \times 1}
\end{align}
we omit the mean subtraction for simplicity as it doesn't affect the general gradient analysis.

Denote the loss as $\mathcal{L}$, for a general gradient flow for a representation $h_i$ from unique source, the model specific parameters (the learnable parameters $w_{r}$ for message passing framework with r as the specific relation) can be computed as:
\begin{align}
    \frac{\partial \mathcal{L}}{\partial w_{r}} = \frac{\partial \mathcal{L}}{\partial h_{f}} \frac{\partial h_{f}}{\partial h_{i}}
    \frac{\partial h_{i}}{\partial w_{r}} \label{eq:gradienttotal}
\end{align}
$\frac{\partial \mathcal{L}}{\partial h_{f}}$ depends on the loss function and $\frac{\partial h_{i}}{\partial w_{r}}$ depends on specific message passing architecture. The intermediate gradient $ \frac{\partial h_{f}}{\partial h_{i}}$ is:
\begin{align}
        \frac{\partial h_{f}}{\partial h_{i}} &= \tilde{\theta}_{i}+\sum_{j}h_{j}\frac{\partial \tilde{\theta}_{j}}{\partial h_{i}}+\frac{1}{k} \label{eq:gradient} \\ 
        &= \tilde{\theta}_{i}+\frac{1}{k}\nonumber\\&+(\frac{1}{\theta_{\uparrow}-\theta_{\downarrow}}-\frac{\theta_{j}-\theta_{\downarrow}}{(\theta_{\uparrow}-\theta_{\downarrow})^2}-\frac{\theta_{\uparrow}-\theta_{j}}{(\theta_{\uparrow}-\theta_{\downarrow})^2}) \nonumber\\
        &\sum_{j}h_{j}(\mathrm{diag}(W^{'}_{ij})W_{i}^T)
\end{align}
From Eq.~(\ref{eq:gradient}), we observe that the gradient for representations from each unique sources is directly related to two factors: its normalized attention score and the raw attention difference.
\begin{theorem}\label{theorem:residual}
     Assuming for a common loss function $\mathcal{L}$ and message passing framework, $\frac{\partial \mathcal{L}}{\partial h_{f}}$ and $\frac{\partial h_{i}}{\partial w_{r}}$ are bounded, further assuming that node representation $h$, learnable parameters $W'$,$W_{i}$ are bounded, then $\frac{\partial \mathcal{L}}{\partial w_{r}}\rightarrow 0$ for extreme low attention $\tilde{\theta}_{i}$ and high raw attention difference $(\theta_{\uparrow}-\theta_{\downarrow})$ without residual connection. Adding a residual coefficient $\frac{1}{k}$ ensures that $ \frac{\partial \mathcal{L}}{\partial w_{r}}$ is lower bounded without converging to 0.
\end{theorem}
\begin{proof}
    As we have shown in Eq.~(\ref{eq:gradienttotal}), assuming $\frac{\partial \mathcal{L}}{\partial h_{f}}$ and $\frac{\partial h_{i}}{\partial w_{r}}$ are bounded, total gradient would vanish only if the intermediate gradient in Eq.~(\ref{eq:gradient}) vanishes to 0. Without residual coefficient,
    \begin{align} \label{eq:intermediategradient}
        \frac{\partial h_{f}}{\partial h_{i}} &= \tilde{\theta}_{i}+(\frac{1}{\theta_{\uparrow}-\theta_{\downarrow}}-\frac{\theta_{j}-\theta_{\downarrow}}{(\theta_{\uparrow}-\theta_{\downarrow})^2}-\frac{\theta_{\uparrow}-\theta_{j}}{(\theta_{\uparrow}-\theta_{\downarrow})^2}) \nonumber\\
        *&\sum_{j}h_{j}(\mathrm{diag}(W^{'}_{ij})W_{i}^T)
    \end{align}
    heavily depends on the attention score learned. 
    
    Since $\tilde{\theta}$ is minmax normalized, the min attention score would make $\tilde{\theta}_{i}$ always be 0.
    
    Assume that $\forall{i,j}$, $h_j$, $W'$, and $W_i$ are bounded, a high difference between maximum and minimum raw attention would lead the second term in Eq.~(\ref{eq:intermediategradient}) to vanish. The total gradient $\frac{\partial \mathcal{L}}{\partial w_{r}}$ would be 0 following the intermediate gradient to be approximately 0.
    Adding a residual coefficient $\frac{1}{k}$, ensures that the intermediate gradient for min attention score is at least $\frac{1}{k}$ and therefore avoid the total gradient $\frac{\partial \mathcal{L}}{\partial w_{r}}$ to vanish.
\end{proof}

\begin{remark}
Theorem~\ref{theorem:residual} necessities the residual connection in terms of alleviating gradient vanishing problem occurred for low attention node representation learning with high raw attention difference. Without residual connection, base models with low attention scores in ensemble learning cannot obtain useful backward gradient updates, leading to inferior performance. Ensemble models cannot absorb unique information from diverse paths. Adding a constant residual coefficient guarantees normal gradient updates for all base models and therefore encourages all base models to learn from final loss.
\end{remark} 

\section{Experiment}
\begin{table*}[htbp]
\caption{Benchmark Dataset Data Statistics}
\centering
{%
\begin{tabular}{llrll}
\hline
Dataset  & Node Type (\# of nodes)               & \multicolumn{1}{l}{Features} & Edge Type (\# of edges)                                         & \# of Labels                   \\ \hline
DBLP~\cite{citation}     & P:Paper(14,328)          & 334                          & PA(19,645)                                         & \multicolumn{1}{r}{4}   \\
         & A:Author(4,057)          & 4,231                         & PT(85,810)                                         &                         \\
         & T:Term(7,723)            & 50                           & PC(14,328)                                         &                         \\
         & C:Conference(20)        & 128                          &                                                   &                         \\ \hline
IMDB~\cite{mag}     & M:Movie(4,278)           & 3,066                         & MA(12,828)                                         & \multicolumn{1}{r}{3}   \\
         & A:Actor(5,257)           & 3066                         & MD(4,278)                                          &                         \\
         & D:Director(2,081)        & 3,066                         &                                                   &                         \\ \hline
ACM~\cite{citation}      & P:Paper(3,025)           & 1,902                         & PA(9,949)                                          & \multicolumn{1}{r}{3}   \\
         & A:Author(5,959)          & 1,902                         & PS(3,025)                                          &                         \\
         & S:Subject(56)           & 1,902                         & PT(255,619)                                        &                         \\
         & T:Term()                & 128                          & PP(5,343)                                          &                         \\ \hline
Freebase~\cite{freebase} & B:Book(40,402)           & 128                          & BB(270,106),BF(38,299),BS(6,615),BL(26921),BO(21,900) & \multicolumn{1}{r}{8}   \\
         & F:Film(19,427)           & 128                          & FF(87,838),MB(31,486),MF(11,291),MM(283,670),MS(8,975) &                         \\
         & M:Music(82,351)          & 128                          & ML(42,915),SF(6,763),SS(1,290),SL(656),PB(35,587)     &                         \\
         & S:Sports(1,025)          & 128                          & PF(17,604),PM(10,948),PS(14,850),PP(22,813),PL(15,134) &                         \\
         & P:People(17,641)         & 128                          & PO(2,215),PBu(5,378),LF(21,299),LL(47,817),OF(13,128)  &                         \\
         & L:Location(9,368)        & 128                          & OM(10,702),OS(559),OL(2,696),OO(1,101),OBu(1,073)     &                         \\
         & O:organization(2,731)    & 128                          & BuB(18,625),BuF(8,397),BuM(24,764)                   &                         \\
         & Bu:Business(7,153)       & 128                          & BuS(610),BuL(6,647),BuBu(4,448)                     &                         \\ \hline
Ogb\_Mag~\cite{ogb} & P:Paper(736,389)         & 128                          & AI(1,043,998)                                       & \multicolumn{1}{r}{349} \\
         & A:Author(1,134,649)       & 128                          & AP(7,145,660)                                       &                         \\
         & I:Institution(8,740)     & 128                          & PP(5,416,271)                                       &                         \\
         & F:Field of Study(59,965) & 128                          & PF(7,505,078)                                       &                        \\ \hline
\end{tabular}%
}
\label{tab:dataset}
\end{table*}

\subsection{Benchmark Datasets}
Five real-world heterogeneous graphs are used as our benchmark datasets. Table~\ref{tab:dataset} summarizes their statistics.

\noindent \textbf{DBLP:}
    We use the DBLP dataset~\cite{citation}, which includes 4,057 authors (A), 14,328 papers (P), 7,723 terms (T), and 20 conferences (C). Authors are categorized into four research fields: database, data mining, artificial intelligence, and information retrieval. Each author is represented using a bag-of-words vector derived from the keywords of their papers. For semi-supervised learning, the author nodes are split into training, validation, and test sets with 400 (9.86\%), 400 (9.86\%), and 3,257 (80.28\%) nodes, respectively. Since conference nodes lack input features, we initialize them with 128-dimensional vectors sampled from a uniform distribution.

\noindent\textbf{IMDB:}
    We evaluate our framework on the IMDB dataset~\cite{mag}, a standard benchmark for graph-based learning. The dataset comprises 4,278 movies (M), 2,081 directors (D), and 5,257 actors (A). Each movie is assigned to one of three genres: Action, Comedy, or Drama. Movies are described using bag-of-words vectors extracted from plot keywords, serving as their input features. For semi-supervised learning, the movie nodes are split into training, validation, and test sets with 400 (9.35\%), 400 (9.35\%), and 3,478 (81.30\%) nodes, respectively.

\noindent \textbf{ACM:}
    We use the ACM dataset~\cite{citation}, which includes 3,025 papers (P), 5,835 authors (A), and 56 subjects (S). Papers are categorized into three research areas: Database, Wireless Communication, and Data Mining. Each paper is represented using a bag-of-words vector based on its keywords. For semi-supervised learning, the author nodes are split into training, validation, and test sets, containing 635 (21\%), 272 (9\%), and 2,118 (70\%) nodes, respectively. 

\noindent\textbf{Freebase:}
    We use the Freebase dataset~\cite{freebase}, a large-scale knowledge graph comprising 8 genres of entities. It includes 40,402 books (B), 19,427 films (F), 82,351 music (M), 1,025 sports (S), 17,641 people (P), 9,368 locations (L), 2,731 organizations (O), and 7,153 businesses (Bu). For semi-supervised learning, the book nodes are divided into training, validation, and test sets with 1,671 (22\%), 415 (5\%), and 5,568 (73\%) nodes, respectively. As the dataset lacks node features, we assign each node a 128-dimensional feature vector initialized from a uniform distribution.

\noindent\textbf{Ogb-mag:}
    We use the Ogb-mag dataset~\cite{ogb}, a heterogeneous network composed of a subset of the Microsoft Academic Graph. It includes 736,389 papers (P), 1,134,649 authors (A), 8,740 institutions (I), and 59,965 field of study (F). For semi-supervised learning, the paper nodes are divided into training, validation, and test sets with 629,571 (85\%), 64,879 (9\%), and 41,939 (6\%) nodes, respectively. We pre-processes the original dataset by adding structural features, using ``metapath2vec'', to nodes without any feature.

\subsection{Baselines}
We compare our \method\  with some state-of-art baselines

\noindent\textbf{HAN}~\cite{HAN} is a heterogeneous graph neural network. It uses both node level and semantic level attention to learn node embeddings from different meta-paths.

\noindent \textbf{Ensemble-GNN}~\cite{ensemble5} is a variant of the state-of-the-art ensemble learning method for homogeneous graphs, which combines predictions from multiple GNNs through voting. Our extend this method by using meta-paths to transform a heterogeneous graph into multiple homogeneous graphs, then training diverse base learners on each meta-path-induced graph. Predictions from all base learners across different meta-paths are aggregated through a voting mechanism to produce the final output.

\noindent\textbf{Cluster-GCN}~\cite{cluster} partitions a homogeneous graph into clusters and performs training on these subgraphs to preserve graph locality and reduce memory usage. To adapt Cluster-GCN for heterogeneous graphs, we convert a heterogeneous dataset into homogeneous by unifying node features across types and aligning all nodes into a single feature matrix. Each resulting homogeneous graph is used as input to Cluster-GCN, leveraging its efficient mini-batch training of base learners.

\noindent\textbf{RGCN}~\cite{RGCN} extends traditional GCNs to handle heterogeneous graphs by introducing relation-specific transformations for each edge type, allowing it to model multi-relational data effectively.

\noindent\textbf{SeHGNN}~\cite{SeH} is a heterogeneous graph representation learning that precomputes neighbor aggregation using a lightweight mean aggregator and avoids repeated computations during training. SeHGNN extends the receptive field with long meta-paths and fuses features through a transformer-based module.

\noindent\textbf{NaiveWeighting-GNN} is a simplified variant of the proposed \method. It retains the same architecture, loss function, and regularization as \method. It only replaces the residual-attention fusion mechanism with a simple voting mechanism that aggregates all base learners outputs. Comparing this variant to \method\ highlights the benefits of using residual-attention fusion over naive averaging.

\subsection{Implementation Details}

We begin with the full set of relations in the heterogeneous graph and form candidate relation groups. Each candidate group is evaluated by training the base model and measuring validation accuracy, with the relation groups that have the highest validation performance selected. 
We conduct a grid search over a selected range of hyperparameters, including hidden dimension: [32,64,128], number of layers: [2,3], feature dropout rate: [0,0.1,0.2], and number of neighbors: [10,15,20]. Adam~\cite{adam} is used as the optimizer. The learning rate, weight decay, regularization rate, and number of training epochs are fixed, with early stopping applied. For each method, we report the average accuracy over five different random seeds. All experiments are performed on desktop workstations equipped with NVIDIA RTX A6000 Ada Generation GPUs.

\subsection{Results and Analysis}
\begin{table*}[htbp]
\centering
\caption{Performance comparisons between baselines and our proposed method across five heterogeneous datasets. Accuracies (ACC) are reported over 5 different initialization status. Superscript *  indicates that \method\ is statistically significantly better than this method at 95\% confidence level using the performance metrics.}
{%
\begin{tabular}{llllll}
\hline
Model & \multicolumn{1}{c}{\begin{tabular}[c]{@{}c@{}}DBLP\\ Accuracy\end{tabular}} & \multicolumn{1}{c}{\begin{tabular}[c]{@{}c@{}}IMDB\\ Accuracy\end{tabular}} & \multicolumn{1}{c}{\begin{tabular}[c]{@{}c@{}}ACM\\ Accuracy\end{tabular}} & \multicolumn{1}{c}{\begin{tabular}[c]{@{}c@{}}Freebase\\ Accuracy\end{tabular}} & \multicolumn{1}{c}{\begin{tabular}[c]{@{}c@{}}Ogb-mag\\ Accuracy\end{tabular}} \\ \hline
HAN & 0.885$_{\pm0.0097}^*$ & 0.555$_{\pm0.0192}^*$ & 0.864$_{\pm0.0047}^*$ & 0.488$_{\pm0.0068}^*$ & 0.251$_{\pm0.0009}^*$ \\
Ensemble-GNN & 0.935$_{\pm0.0017}^*$ & 0.580$_{\pm0.0043}^*$ & 0.910$_{\pm0.0094}$ & 0.536$_{\pm0.0059}^*$ & 0.412$_{\pm0.0047}^*$ \\
Cluster-GNN & 0.800$_{\pm0.0075}^*$ & 0.531$_{\pm0.0038}^*$ & 0.836$_{\pm0.0133}^*$ & 0.535$_{\pm0.0037}^*$ & 0.307$_{\pm0.0042}^*$ \\
RGCN & 0.935$_{\pm0.0009}^*$ & 0.568$_{\pm0.0018}^*$ & 0.899$_{\pm0.0019}^*$ & 0.541$_{\pm0.0320}^*$ & 0.406$_{\pm0.0039}^*$ \\
SeHGNN & 0.878$_{\pm0.0094}^*$ & 0.583$_{\pm0.0034}$ & 0.753$_{\pm0.0213}^*$ & 0.492$_{\pm0.0019}^*$ & 0.443$_{\pm0.0009}^*$ \\
NaiveWeighting-GNN & 0.942$_{\pm0.0019}^*$ & 0.563$_{\pm0.0060}^*$ & 0.897$_{\pm0.0073}^*$ & 0.511$_{\pm0.0048}^*$ & 0.344$_{\pm0.0265}^*$ \\
LHGEL & \textbf{0.950}$_{\pm0.0013}$ & \textbf{0.588}$_{\pm0.0015}$ & \textbf{0.912}$_{\pm0.0099}$ & \textbf{0.578}$_{\pm0.0089}$ & \textbf{0.460}$_{\pm0.0081}$ \\ \hline
\end{tabular}%
}
\label{tab:results}
\end{table*}

\paragraph{Baseline Comparison}
Table~\ref{tab:results} reports the results across five heterogeneous graph datasets where \method\, demonstrates strong performance gains compared to baseline models. Under the same message passing scheme, \method\ achieves statistically significant improvements (with 95\% confidence) on four datasets-DBLP, IMDB, Freebase, and Ogb-mag. On the ACM dataset, \method\ performs comparably with Ensemble-GNN, achieving the highest accuracy, highlighting its robustness across various graph structures and node types.

Notably, \method\ outperforms Ensemble-GNN with statistical significance on 4 out of 5 datasets. While both approaches fall under the graph ensemble learning paradigm, Ensemble-GNN lacks a unified objective to guide its base learners and does not impose constraints to promote diversity among them. As a result, its base models tend to be less accurate and more redundant, leading to weaker overall performance compared to the more coordinated and diversity-aware design of \method.

When comparing \method\ with its variant NaiveWeighting-GNN, results demonstrate that \method\ consistently and significantly outperforms the variant across all five datasets, highlighting the effectiveness of the proposed residual-attention mechanism over simple averaging. These findings also align well with our theoretical analysis.

\paragraph{Abalation Study on Residual-Attention \& Regularizer}
Table~\ref{tab:ablation} presents the ablation results of \method\ using two residual-attention mechanism (minmax/softmax) in conjunction with the regularizer, evaluated across five benchmark datasets. Notably, the minmax-based residual attention consistently outperforms its softmax counterpart on all datasets, demonstrating a stronger capacity to discriminate the relative importance of base learners during aggregation. The regularizer further enhances performance by enforcing predictive consistency across learners, which contributes to improved generalization and robustness. These findings highlight the complementary strengths of the framework in handling large heterogeneous graphs. In particular, the Freebase dataset exhibits substantial gains when both the regularizer and minmax attention are applied, underscoring their effectiveness in settings with high structural heterogeneity and feature noise.
\begin{table}[htbp]
\caption{Ablation study results \textit{w.r.t.} regularizer and residual-attention mechanism}
\centering
{%
\begin{tabular}{lll}
\hline
Dataset  & Model                    & Accuracy            \\ \hline
DBLP     & LHGEL minmax+regularizer & \textbf{0.950}$_{\pm0.0013}$ \\
         & LHGEL softmax            & 0.943$_{\pm0.0022}$ \\
         & LHGEL minmax             & 0.943$_{\pm0.0039}$ \\ \hline
IMDB     & LHGEL minmax+regularizer & \textbf{0.598}$_{\pm0.0031}$ \\
         & LHGEL softmax            & 0.568$_{\pm0.0131}$ \\
         & LHGEL minmax             & 0.570$_{\pm0.0186}$ \\ \hline
ACM      & LHGEL minmax+regularizer & \textbf{0.921}$_{\pm0.0145}$ \\
         & LHGEL softmax            & 0.892$_{\pm0.0201}$ \\
         & LHGEL minmax             & 0.887$_{\pm0.0185}$ \\ \hline
Freebase & LHGEL minmax+regularizer & \textbf{0.578}$_{\pm0.0089}$ \\
         & LHGEL softmax            & 0.535$_{\pm0.0134}$ \\
         & LHGEL minmax             & 0.536$_{\pm0.0195}$ \\ \hline
Ogb-mag  & LHGEL minmax+regularizer & \textbf{0.460}$_{\pm0.0081}$ \\
         & LHGEL softmax            & 0.444$_{\pm0.0074}$ \\
         & LHGEL minmax             & 0.440$_{\pm0.0110}$ \\ \hline
\end{tabular}%
}
\label{tab:ablation}
\end{table}

\begin{figure*}[htbp]
    \centering
    \includegraphics[width=1\textwidth]{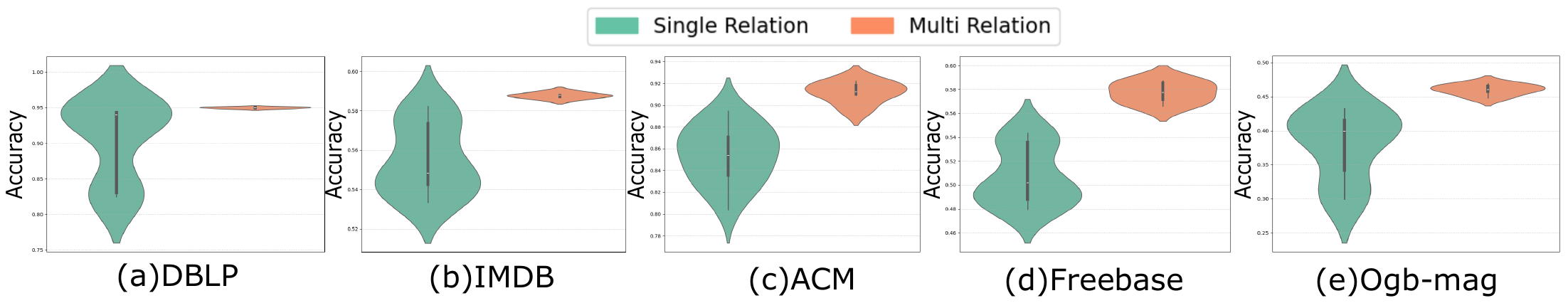}
    \caption{Impact of the number of relation groups on the ensemble learning results. Green violin plots show mean and variance for \method\ with only single relation group, whereas orange violin plots show \method's mean and variance with multiple relation groups.}
    \label{fig:metapath}
\end{figure*}

\paragraph{Abalation Study on Relation Groups \& Batch Sizes}
To evaluate the effect of relation group diversity on model performance, we conduct an ablation study by isolating a single relation group while keeping all other training settings unchanged, including the use of multiple batch sizes. As shown in Figure~\ref{fig:metapath}, we compare the mean and variance of accuracies across different numbers of relation groups. The green violin plots represent results for \method\ with a single relation group, while the orange plots correspond to \method\ using multiple relation groups. For each dataset, results for the multi-relation relation setting are shown directly after the single setting. Notably, the green plots exhibit substantially higher variance and lower mean accuracy compared to their multi-relation counterparts. These findings highlight the importance of relation group diversity in stabilizing training and enhancing generalization in our model.

\begin{figure*}[htbp]
    \centering
    \includegraphics[width=1\textwidth]{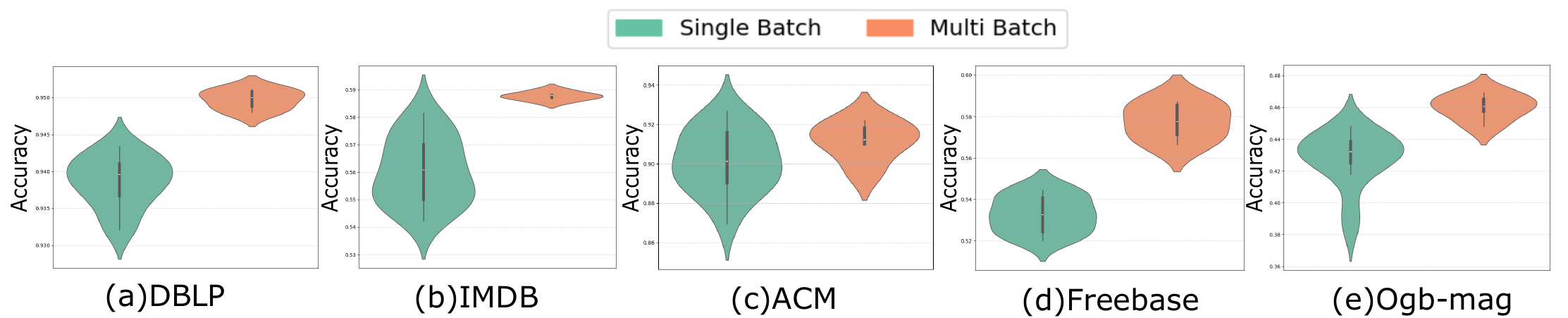}
    \caption{Impact of the number of batch sizes on the ensemble learning results. Green violin plots show mean and variance for \method\ with only single batch size, whereas orange violin plots show \method's mean and variance with multiple batch sizes.}
    \label{fig:batch}
\end{figure*}

We also explored how batch size diversity affects training. We trained models using a single batch size (represented by green violin plots in Figure~\ref{fig:batch} while keeping all other conditions constant, including multiple relation groups. For comparison, orange plots show results from multiple batch sizes.
Our findings indicate that configurations with diverse batch sizes significantly outperform their single-batch counterparts. The green plots (single batch size) show higher variance on four of the datasets and lower accuracy on all datasets. This highlights that diverse batch sizes contribute to more stable training and improved generalization of the model.
\begin{figure}[htbp]
    \centering
    \includegraphics[width=0.6\linewidth]{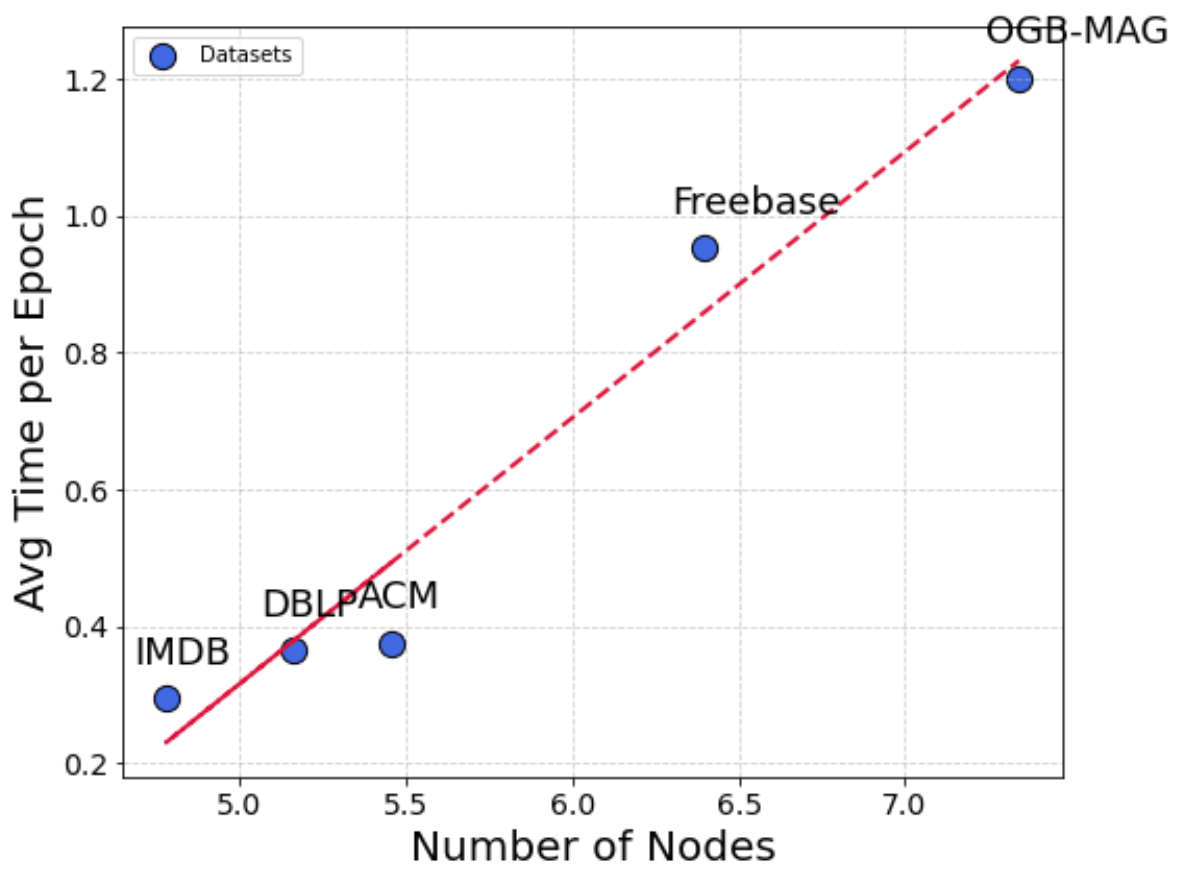}
    \caption{Log average runtime per epoch ($y$-axis) using fixed batch sizes \textit{vs.} the sum of number of nodes and number of edges in log scale ($x$-axis).}
    \label{fig:time}
\end{figure}

\paragraph{Complexity analysis}
We report the wall-clock runtime performance on five benchmark datasets to demonstrate the scalability of our method. As shown in Figure~\ref{fig:time}, the average training time per epoch exhibits a clear increasing trend as the number of nodes and edges grows. Notably, this growth follows a linear pattern on a log-log scale, indicating that our method scales efficiently with graph size, even on large heterogeneous graphs.

\section{Conclusion}
This paper introduced \method, a novel ensemble learning framework tailored for large heterogeneous graphs. Unlike prior ensemble methods mainly designed under homogeneous or IID settings, \method\ directly addresses the challenges posed by graph heterogeneity, including diverse node and edge types and and complex local structures.
\method\ strives to train accurate and diverse base classifiers by leveraging a regularized ensemble framework that operates on subgraphs sampled in varying batch sizes. 
%
To mitigate bias introduced by batch sampling, \method\ incorporates a batch view aggregation mechanism that enables information fusion across different subgraph views.
%
A key novelty of our framework is its two-stage residual attention mechanism, which firstly  aggregates embeddings from different batch sizes within each relation group, and then integrates embeddings across relation groups. This hierarchical attention structure enables adaptive ensemble weighting at multiple levels, promoting both predictive accuracy and robustness.
Our theoretical analysis establishes the convergence properties of \method\ and its advantages over na\"ive ensemble baselines. Extensive experiments on five benchmark heterogeneous graph datasets demonstrate that \method\ consistently outperforms the state-of-the-art methods in accuracy, stability, and robustness. 

\section*{Acknowledgment}
This work has been supported in part by the National Science Foundation (NSF) under
Grant Nos.\, IIS-2505719, IIS-2236579, IIS-2302786, IIS-2441449, IOS-2430224, and IOS-2446522,
the Science Center for Marine Fisheries (SCeMFiS), and the Commonwealth Cyber Initiative (CCI).


\bibliographystyle{ieeetr}
\bibliography{ICDM25}

\end{document}